\documentclass[10pt,twocolumn,letterpaper]{article}

\usepackage{cvpr}              
\usepackage{tabularx}
\usepackage{ragged2e}

\newcolumntype{L}{>{\RaggedRight\arraybackslash}X}

\usepackage{url}
\usepackage[utf8]{inputenc} 
\usepackage[T1]{fontenc}    
\usepackage{booktabs}       
\usepackage{amsfonts}       
\usepackage{nicefrac}       
\usepackage{microtype}      
\usepackage{makecell}
\usepackage{setspace}
\usepackage{multirow}
\usepackage{graphicx}
\usepackage{subcaption}
\usepackage{pifont}%
\usepackage{xspace}
\usepackage{float}
\usepackage{mdframed}
\usepackage{fancyvrb}
\usepackage{listings}

\providetoggle{showcomments}
\settoggle{showcomments}{true} 

\iftoggle{showcomments}{
    \newcommand{\david}[1]{{\color{orange}[DR: #1]}}
    \newcommand{\arsha}[1]{{\color{blue}[Arsha: #1]}}
    \newcommand{\jasper}[1]{{\color{violet}[Jasper: #1]}}
    \newcommand{\cs}[1]{{\color{OliveGreen}[Cordelia]: #1]}}
    \newcommand{\tobias}[1]{{\color{RoyalPurple}[Tobias]: #1]}}
    
}{
    \newcommand{\david}[1][]{}    
    \newcommand{\arsha}[1]{}
    \newcommand{\jasper}[1][]{}
    \newcommand{\cs}[1][]{}    
    \newcommand{\tobias}[1][]{}
    
}

\newcommand{\para}[1]{\noindent\textbf{#1}}
\floatstyle{plaintop}
\restylefloat{table}
\usepackage[export]{adjustbox}
\usepackage{mdframed}
\newcommand{\dataset}{Minerva-Ego}

\DeclareCaptionLabelFormat{andtable}{#1~#2  \&  \tablename~\thetable}

\definecolor{cvprblue}{rgb}{0.21,0.49,0.74}
\usepackage[pagebackref,breaklinks,colorlinks,allcolors=cvprblue]{hyperref}

\def\paperID{} 
\def\confName{CVPR}
\def\confYear{2026}

\title{\dataset{}: Spatiotemporal Hints for Egocentric Video Understanding}

\author{
Arsha Nagrani$^{*}$
\and Jasper Uijlings$^{*}$ \and Shyamal Buch$\thanks{Equal contribution.  Correspondence to {\tt anagrani@google.com}}$ \and Tobias Weyand \and Sudheendra Vijayanarasimhan \and Bo Hu \and Ramin Mehran \and David A Ross \and Cordelia Schmid\\
 \\
Google DeepMind \\
}

\begin{document}

\newmdenv[backgroundcolor=gray!10, linecolor=black, linewidth=0.5pt, roundcorner=4pt, innerleftmargin=2pt, innerrightmargin=2pt, innertopmargin=5pt, innerbottommargin=5pt, font=\scriptsize\ttfamily]{exampleprompt}

\maketitle
\begin{abstract}
Video reasoning models are a core component of egocentric and embodied agents. However, standard benchmarks for assessing models provide only evaluation of the output (\eg the answer to a question), without evaluation of intermediate reasoning steps, and most provide answers only in the text domain. 
We introduce \dataset{}, a benchmark for evaluating complex egocentric visual reasoning. We extend recent high-quality video data sources recorded from egocentric / embodied settings with a set of challenging, multi-step multimodal questions and spatiotemporally-dense human-annotated reasoning traces. Benchmarking experiments show that state-of-the-art models still have a large gap to human performance. To investigate this gap in detail, we annotate each reasoning trace in the dataset with the objects of interest required to solve the question, as spatio-temporal mask annotations. Through extensive evaluations, we identify that prompting frontier models with hints of `where' and `when' to look yields substantial improvements in performance. \dataset{} can be downloaded at \url{https://github.com/google-deepmind/neptune}

\end{abstract}
\section{Introduction}
\label{sec:intro}
\begin{figure*}
\vspace{-.7cm}
\centering
  \includegraphics[width=0.95\linewidth]{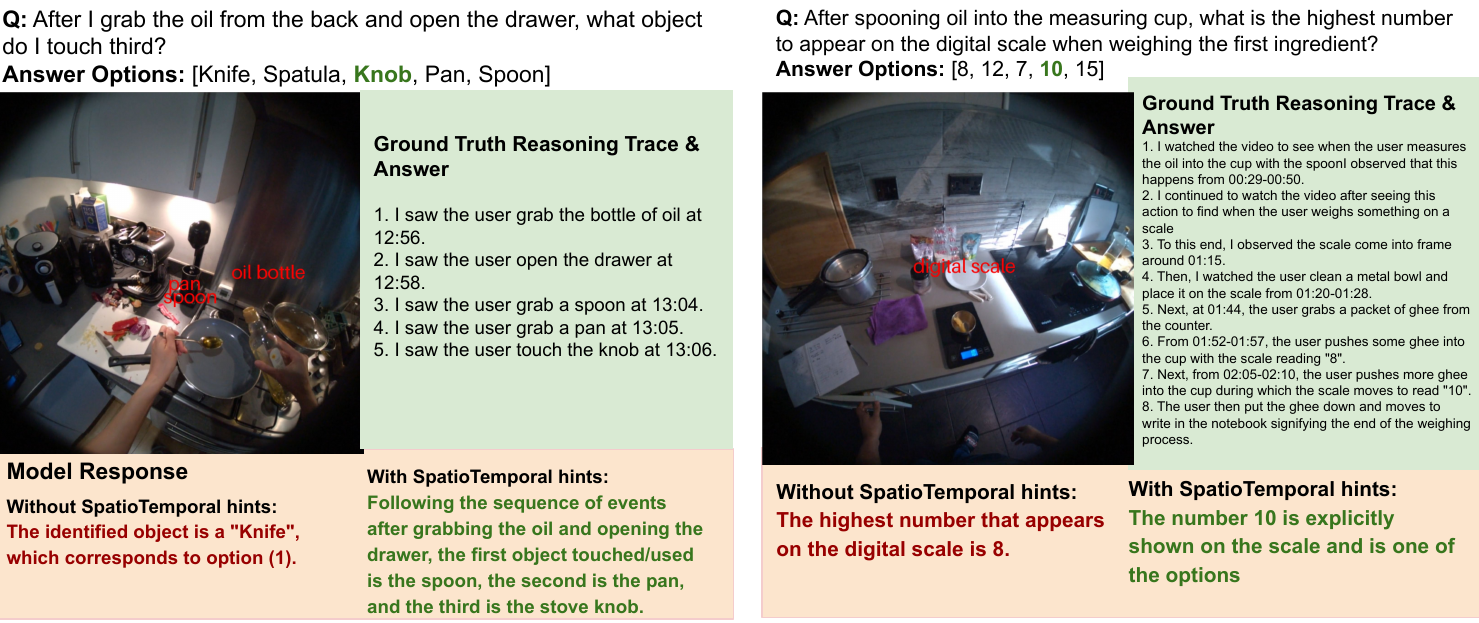}
  \caption{\textbf{Spatio-temporal hinting in \dataset{}:} We introduce \dataset{}, a complex video question-answering dataset for egocentric videos. Unlike prior work \citep{plizzari2024egptempo,Chandrasegaran2024hourvideo,mangalam2023egoschema,perrett2025hdepic}, the answer to each question is accompanied by a spatiotemporally-grounded \textit{reasoning trace}, which outlines the steps required to come to the answer which are grounded in time (localization with timestamps) and space (associations with segmentation masks of objects). We find SOTA models run naively tend to make reasoning errors, but by simply prompting models `where' and `when' to look via spatiotemporal hints, we can circumvent these errors.
  }
  \vspace{-2mm}
  \label{fig:teaser}
  \end{figure*}

Understanding and reasoning over events in the visual world, both fine-grained and over long temporal spans, is a core capability of humans. Reasoning about videos from an egocentric perspective is foundational to enabling effective and broad deployment of assistive and embodied agents~\cite{feng2025embodied,liu2023egocentric}.

Broadly, there has been remarkable progress towards the development of video reasoning methods, evolving from simple action recognition on short clips~\citep{yu2019activitynet,kay2017kinetics,goyal2017something} to more advanced capabilities over longer time horizons (LVBench~\citep{wang2024lvbench}, VideoMME~\citep{fu2024video}, Neptune~\citep{nagrani2024neptune}, and others). The development of the latter has been accelerated by the rise of multimodal LLMs~\citep{achiam2023gpt, reid2024gemini, bai2025qwen2}, which can perform complex video tasks on hour long videos~\citep{wang2024lvbench}, and are steadily approaching human performance on these recent benchmarks~\citep{nagrani2024neptune,fu2024video,nagrani2025minerva}.

However, despite this progress, a critical gap remains in evaluating complex reasoning for long-form, egocentric video. Existing egocentric benchmarks often focus solely on the final answer and overlook the underlying reasoning process~\cite{mangalam2023egoschema}. While recent work such as EgoTempo has begun to address temporal localization~\cite{plizzari2024egptempo}, it fails to assess reasoning over extended contexts. 

Attempts to improve the reasoning process of models have led to a paradigm shift where models spend more time \emph{thinking}~\cite{openai-reasoning} i.e.\ producing intermediate thoughts towards a final answer, often via additional compute at inference time~\cite{jones2021scaling,snell2024scaling}. 
The idea has been introduced in the video domain as well, where benchmarks such as MINERVA~\citep{nagrani2025minerva} assess the types of errors made specifically for video reasoning. These works, however are primarily exocentric. Unlike exocentric videos, which often feature detached observations, egocentric data centers on close-up, frequent interactions with objects, producing a continuous flow of activities that demand a model's ability to identify and focus on key objects of interest over time.


  
To address these limitations, we introduce a new benchmark for long-form egocentric video understanding, called \dataset{}.
Our benchmark, built on top of HD-EPIC~\cite{perrett2025hdepic}, features 1,160 complex, hand-crafted questions that require reasoning over long temporal contexts. Crucially, each question includes five answer choices annotated with dense spatio-temporal reasoning traces, which link timestamps and segmentation masks to relevant objects of interest (see Fig.~\ref{fig:teaser} for examples).

We find that state-of-the-art video understanding models (Gemini~\cite{gemini2.5}, GPT~\cite{openai25gpt5} and Claude~\cite{claude35sonnetv2}) perform poorly on our dataset, with a large gap to human performance. Analysing the errors in more detail, we find that models make reasoning mistakes in two key areas---perceptual identification and temporal grounding of objects---which is especially significant since object manipulation is central to egocentric video.
To remedy this, we perform experiments pointing models towards `where' and `when' to look. We do this by employing both oracle and predicted segmentations to selectively emphasize relevant objects via visual highlighting on the frames directly. In oracle experiments using ground-truth and SAM2-interpolated tracks, we found that encircling objects and displaying class names, and temporal selection—which focuses input frames on relevant object occurrences—yielded a peak improvement of 5.6\%. When replicating this with an open-vocabulary segmentation model (OWLv2~\cite{minderer23neurips_owlv2}), temporal selection and class name visualization continued to provide modest gains.

To summarize, we introduce \dataset{}, the first egocentric benchmark to pair complex hand-crafted multiple choice questions with dense spatiotemporally-grounded reasoning traces. This unique annotation structure allows us to move beyond simple accuracy metrics to diagnosable evaluation, proving that current models fail primarily due to perceptual grounding errors—a finding we validate through our spatiotemporal highlighting experiments.

\section{Related Work}
\label{sec:related-work}

\subsection{Recent VideoQA Datasets}
Many valuable video datasets have been proposed recently. TemporalBench~\citep{Cai2024temporalbench} and PerceptionTest~\citep{patraucean2024perception} include a variety of tasks, such as video QA, captioning, and grounding, but use relatively short videos (a couple of minutes). 
Similarly, CinePile~\citep{rawal2024cinepile} and VideoVista~\citep{li2024videovista} focus on short-form content (avg ${\sim}160$s). 
VideoVista~\citep{li2024videovista} is notable for its broad coverage of 19 understanding and 8 reasoning tasks.
These benchmarks, along with InfiniBench~\citep{Ataallah2024infinibench} (max 52 min videos), and Neptune~\citep{nagrani2024neptune}, rely on semi-automatic pipelines using LLMs for annotation. In contrast, our dataset is entirely manually annotated focusing on quality.  
TVBench~\citep{Cores2024tvbench} specifically focuses on temporally challenging questions which require visual information. 
{E}.{T}. bench~\citep{Liu2024etbench} focuses on multi-event and time-sensitive understanding, using human-annotated videos with a maximum length of 795s. 
Video-MME~\citep{fu2024video} bridges the gap between short and long-form content, and datasets such as 
LVBench~\citep{wang2024lvbench} (avg 68 mins)
extend to even longer videos. 
In contrast to these existing benchmarks, our work provides human-annotated reasoning traces with spatio-temporal grounding, which enables evaluating the model's reasoning ability on top of reporting accuracy.

\begin{figure*}[t]
\vspace{-.7cm}
\centering
  \includegraphics[width=0.9\linewidth]{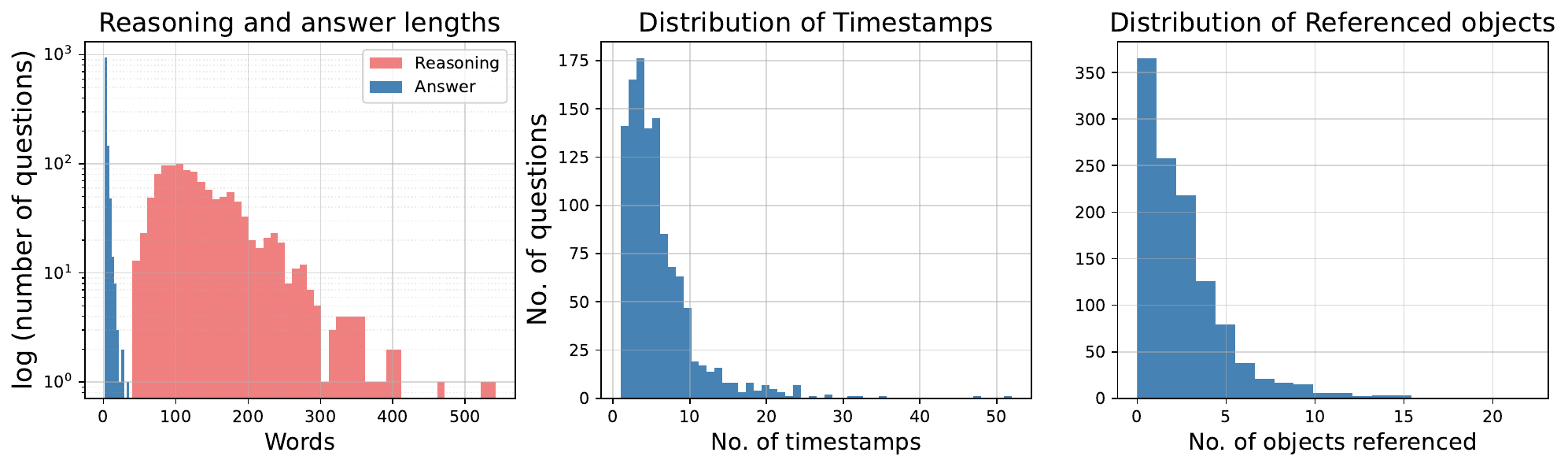}
\centering
\caption{\textbf{Statistics of Reasoning Traces in \dataset{}.} We show lengths of answers and reasoning traces (left), the distribution of timestamps mentioned in the reasoning traces (middle) and the distribution of objects referenced in the reasoning traces (right). Almost all reasoning traces contain at least one timestamp and refer to at least one object, for which a spatio-temporal mask is available. Reasoning traces are long and detailed, and the mean number of timestamps per question is 6.3, while the mean number of objects referenced is 2.8. }
\label{fig:dataset-stats}
\vspace{-.2cm}
\end{figure*}

\subsection{Egocentric Datasets}
A few recent datasets focus specifically on Egocentric VQA. EgoSchema~\cite{mangalam2023egoschema} is a long-form video question-answering dataset containing over 5,000 human curated multiple choice question answer pairs from Ego4D~\cite{grauman2022ego4d} with five choices.
EgoTempo~\cite{plizzari2024egptempo}, a dataset specifically designed to evaluate temporal understanding in the egocentric domain, includes 500 question-answer pairs on egocentric videos with an average duration of 45s.
HourVideo~\cite{Chandrasegaran2024hourvideo} targets very long videos spanning durations of 20-120 mins and features 12,976 five-way
multiple-choice questions from 500 manually curated egocentric videos selected from Ego4D~\cite{grauman2022ego4d}.
HD-EPIC~\cite{perrett2025hdepic} is a kitchen-based egocentric dataset collected  in-the-wild but with detailed annotations matching those in controlled lab environments. 
Similarly, EgoVQA \citep{egovqa} contains 600 question-answer pairs
and covers queries of actions, positions, objects and counting which are relatively broad (\eg “What am I doing?”) and does not capture fine-grained temporal dynamics.
EgoTaskQA~ \citep{egotaskqa} includes questions from over 2,000 egocentric videos covering actions, goals, and beliefs.
EgoMemoria~\citep{ye2025mmego} derives its questions from the narrations provided for Ego4D videos. 
Unlike \dataset{}, which relies on expert human annotators, both EgoTaskQA and EgoMemoria consist of machine-generated questions and answers.
Finally, EgoThink \citep{egothink} comprises of 700 images extracted from 595 videos,
across several categories.
While it offers more comprehensive, fine-grained capabilities, it relies on static images, limiting its ability to capture temporal information. Unlike all these datasets, \dataset{} contains detailed spatio-temporal reasoning traces for each question. 


\subsection{Reasoning Traces for Video}
As discussed in the previous section, few datasets provide auxiliary data beyond final answers. ScienceQA~\citep{saikh2022scienceqa} consists of science questions about images, and provides the accompanying lectures, which when used for training results in increased performance with fewer data points.
Some recent works have aimed to develop automatic labeling pipelines to instead generate noisy auxiliary information at scale.
Visual CoT~\citep{shao2024visual} provides bounding box pseudolabels generated by off-the-shelf models for image VQA, while VideoEspresso~\citep{han2024videoespresso} does something similar for video, constructing a pipeline connecting different frozen models together to label bounding boxes for sparse key frames. VideoCoT~\citep{wang2024videocot} is perhaps the closest to our work, aiming to generate text-based chain of thought for videos; however, we note key differences. The primary goal of VideoCoT is to present a semiautomatic pipeline for training purposes. In contrast we present a fully manually annotated, high-quality dataset for evaluation. The videos in VideoCoT are all taken from Kinetics700~\citep{kay2017kinetics}, making them shorter and limited in domain. Finally, their automated rationales tend to contain substantial information about the video that is unrelated to the particular query, rather than providing specific reasoning for the given question. 

\subsection{Spatio-temporal Highlighting}



Recent advances in VLMs have spurred interest in \emph{visual prompting} techniques like Set-of-Mark (SoM)~\cite{yang2023setofmark}, which demonstrate that overlaying semantic marks (numbers, masks, or bounding boxes) significantly unlocks grounding capabilities in static images. However, applying these concepts to egocentric video introduces unique challenges, namely reasoning over dynamic, long-form content where objects frequently enter and exit the field of view, undergo occlusion, or change state. We extend visual highlighting to the spatiotemporal domain. Distinct from SoM, as well as \cite{bar2022visual,wu2024visual}, which operate on single frames, we introduce spatiotemporal hints that guide the model not only where to look (via spatial highlighting) but also when to look (via temporal frame selection).

Unlike ``Look, Remember, and Reason''~\cite{bhattacharyya2024look}, which implicitly learns grounding by training a specialized encoder on surrogate tasks, our approach uses explicit spatio-temporal highlighting directly on the pixels. This avoids the need for architectural changes or retraining, enabling us to probe perceptual failures in fixed, frontier models like Gemini and GPT.

\section{\dataset{}}
\subsection{Statistics}
\dataset{} consists of 1,160 questions covering 156 unique egocentric videos. Questions are designed to require multi-step reasoning to solve and come with 5 multiple-choice options, with one being the right answer. Additionally, we provide a detailed reasoning trace for each question that explains how to arrive at the correct answer. The reasoning traces contain references to timestamps in the video and references to objects with ground-truth segmentation masks. All annotations are entirely hand-written by highly skilled raters trained for this task. Fig.~\ref{fig:teaser} shows example questions.

The videos are sourced from the HD-EPIC~\cite{perrett2025hdepic} dataset, a high-quality egocentric video dataset. We re-use their ground-truth object segmentation masks. While HD-EPIC also has QA annotations, they are generated using standardized templates automatically and do not focus on multi-step reasoning.
To maintain manageable inference costs for large models, we limit the dataset to approximately 1,000 questions, given that each requires multi-step reasoning and contains detailed traces.


Detailed statistics are provided in Fig.~\ref{fig:dataset-stats}. Videos cover a diverse range of lengths, from 10 seconds to 75 minutes. Reasoning traces are long and detailed, averaging 122 words each, and refer to both timestamps and objects of interest. The average number of objects referenced in a reasoning trace is 2.8, but some refer up to 20 objects. A more-detailed breakdown appears in Section~\ref{sec:additiona-stats} of the supplementary.

\subsection{Dataset Creation Pipeline} 
\label{sec:dataset-creation}
Our pipeline consists of the following 3 steps: \\
\noindent \textbf{1. Selection:} 
We start with all videos from the HD-EPIC~\cite{perrett2025hdepic} dataset, and overlay the existing ground-truth segmentation masks for all objects labeled on to frames of the video. \\
\noindent \textbf{2. Annotation and Verification:} We use highly qualified raters to annotate questions, answers, and reasoning traces. For reasoning traces, raters are asked to reference the key objects of interest needed to answer the question, after being provided all the segmentation masks for the video. We then perform a quality review process, where we ask a separate set of annotators to try and answer the questions, and for the ones that they got wrong, we ask a third set of annotators to look at both sets of annotation and reconcile them, changing the questions, options or reasoning traces if necessary. 
Note that our entire annotation and quality review process is done manually to ensure high quality and avoid model bias. \\ 
\noindent \textbf{3. Post-processing:} Finally, we post-process the dataset by identifying and filtering modality bias (\eg questions answerable with only the text question and answers) following best practices outlined in prior work ~\cite{nagrani2025minerva,hsieh2023sugarcrepe,le2020adversarial,wang2024lvbench}, with more details provided in the supplementary materials, Sec.~\ref{sec:dataset-creation-appendix}. 

\section{Experiments and Analysis}
We perform two main types of experiments on \dataset{}: SOTA evaluation and Spatio-temporal highlighting.\\ 
\noindent\textbf{SOTA evaluation:} First, we benchmark the best video understanding models on \dataset{}. Unlike other works that simply look at the correctness of final answers, we also compare model reasoning traces to the ground truth, and analyze error modes. 
Additionally, we perform ablations examining the effect of the number of input frames and the thinking capability. 

\noindent\textbf{Spatio-temporal Highlighting:} Second, we perform experiments examining the performance with spatio-temporal highlighting. This is done in two ways: (i) in an oracle manner where we provide the model with hints from our extensive spatiotemporal annotations, including object locations, class labels and the set of relevant frames. We analyze how much the model benefits from these hints, which provides interesting insights into their spatiotemporal reasoning abilities in an egocentric setting, and (ii) in a practical setting using off-the-shelf segmentation tools.

\subsection{State-of-the-art evaluation}
\label{sec:sota-eval}

We benchmark \dataset{} on
Gemini 2.5 Pro~\cite{gemini2.5}, Gemini 2.5 Flash~\cite{gemini2.5}, GPT-4.1~\cite{openai-gpt4.1}, GPT-5~\cite{openai25gpt5} and Claude Sonnet 4.0~\cite{claude4opus_sonnet}, as well as leading open-source video model Qwen3\footnote{Results are obtained using the vLLM library (\url{https://docs.vllm.ai/projects/recipes/en/latest/Qwen/Qwen3-VL.html}}~\cite{qwen3techreport}.
We also conduct a human study, performed by 12 raters disjoint from those who did the annotation. All raters had access to the full videos.

For this comparison, we linearly subsample 64 frames from each video and feed it to the models. We do not use any audio or ASR input since the videos do not have any spoken content.

Details of the human study, prompts, and other hyperparameters are in the supplementary material.







\begin{table}[t]
\centering
\resizebox{0.4\textwidth}{!}{
\begin{tabular}{lccr}
\toprule
\textbf{Method}  & \textbf{\# Frames} & \textbf{Thinking} & \textbf{MCQ-Acc. \%} \\
\midrule
Random & - & - & 20.0   \\
\midrule
Qwen-3~\citep{qwen3techreport} & all @1fps & Yes & 29.3 \\
{Claude Sonnet 4~\cite{claude4opus_sonnet}} & 64 & Yes & 30.5 \\ 
{Gemini 2.5 Flash~\citep{gemini2.5}} & 64 & No & 31.7 \\
{Gemini 2.5 Flash~\citep{gemini2.5}} & 64 & Yes & 35.6 \\ 
{Gemini 2.5 Pro~\citep{gemini2.5}} & 64 & Yes & \textbf{40.1} \\ 
GPT-4.1~\cite{openai-gpt4.1} & 64 & No & 30.8 \\ 
{GPT-5~\citep{openai25gpt5}} & 64 & Yes & \underline{37.6} \\
\midrule
{Human performance} & {all} & Yes & 91.8 \\
\bottomrule
\end{tabular}}
\caption{\textbf{State-of-the-art model evaluation on \dataset{}}.  Here we only evaluate final answers with MCQ accuracy.
 Reasoning evaluation is shown in Fig. \ref{fig:reasoning_eval}.
 Due to LLM context size limits, most models are provided a fixed number of frames (64) for each input video, except Qwen which is provided full input videos subsampled to 1 fps. Humans see the original videos at full frame rate.
 GPT-5 was run with "medium" reasoning effort, while GPT-4.1 does not support reasoning.
}
\label{tab:results} 
\end{table}

\subsubsection{Multiple Choice Question (MCQ) Results}
We begin by evaluating final answers only -- these results are shown in Tab.~\ref{tab:results}. Gemini 2.5 Pro leads with a significant margin (+2.3\%) over the second best model, GPT-5.
Gemini 2.5 Flash without thinking performs similarly to GPT-4.1, which does not support reasoning. 
Enabling thinking for Flash results in a significant performance improvement (+3.9\%).
We only report the performance of Claude Sonnet and Gemini 2.5 Pro with thinking because it was not possible to disable thinking for these models.
Notably, the performance of open-source Qwen-3 is almost on par with proprietary models Claude and GPT 4.1.
We conclude that Gemini 2.5 Pro is the best performing model and that thinking helps for the complex questions in our dataset. Based on these results, we do our subsequent ablation studies using the Gemini family.

\subsubsection{Reasoning Analysis}
\label{sec:reasoning_analysis}
\begin{figure*}
    \centering
    \includegraphics[width=0.3\linewidth]{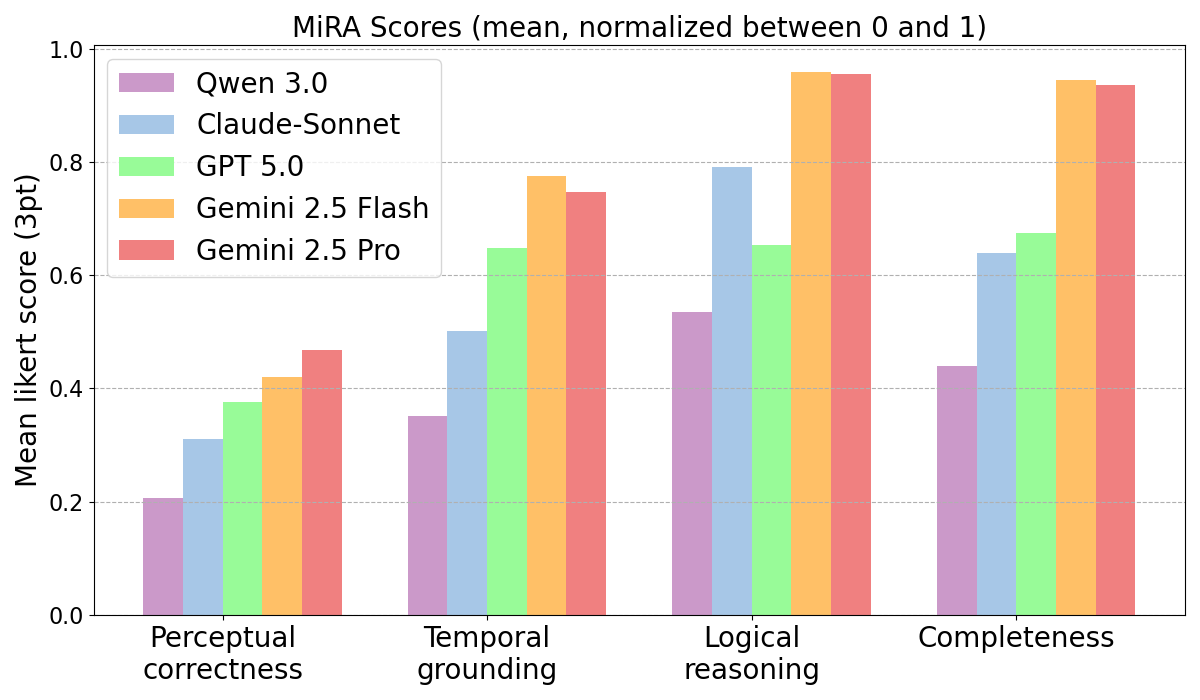} \hspace{0.1in}
     \includegraphics[width=0.3\linewidth]{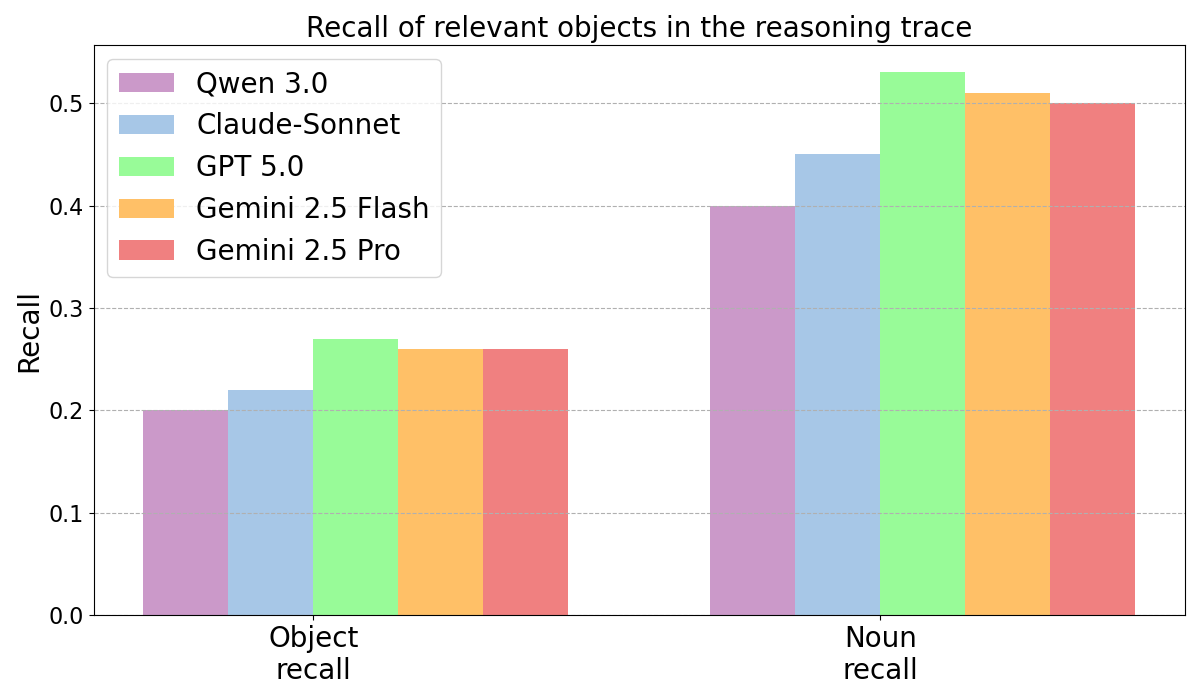}
     \hspace{0.1in}
     \includegraphics[width=0.3\linewidth]{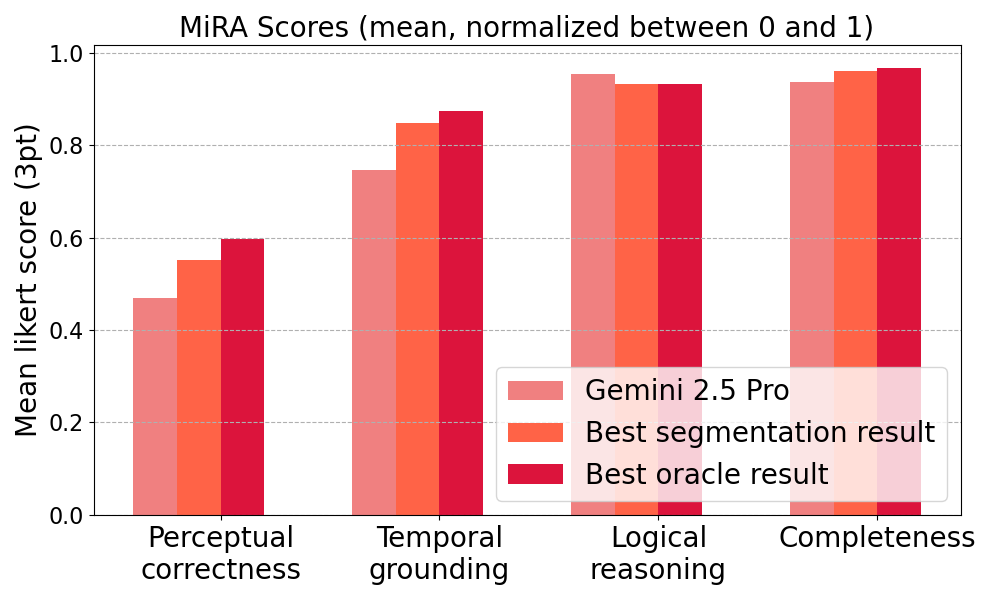}
     \caption{{\bf Analysis of model produced reasoning traces:} (Left) We report MiRA scores on reasoning traces for four axes, namely, Perceptual correctness, Temporal Grounding, Logical reasoning and Completeness. (Middle) We report the recall of relevant objects extracted from the ground truth reasoning trace in the reasoning traces of the models. Object recall matches the entire object's name while Noun recall matches only the main noun in the object's name. (Right) We report the MiRA scores for the best model with oracle segmentation and the best model using real segmentation to show the headroom available for approaches in (Left).}   
\label{fig:reasoning_eval}
\end{figure*}

We evaluated the reasoning traces using an LLM according to the
MINERVA rubric (MiRA) introduced in \cite{nagrani2025minerva}. The MINERVA rubric (repeated in full in suppl. Fig.~\ref{fig:mira_prompt}) evaluates across four dimensions, namely Perceptual correctness, Temporal grounding, Logical reasoning and Completeness, and scores answers on a 3-point Likert scale. Fig.~\ref{fig:reasoning_eval} (left) shows the MiRA scores for four different models. Overall, all four models score well on Logical Reasoning and Completeness while Perceptual correctness is the weakest dimension. On all four dimensions, the Gemini models seem to outperform the others.
To check for scorer bias, we re-evaluated using GPT-5; the relative model ordering is preserved (supplemental Section~\ref{sec:mira-gpt}).

We also evaluated the recall of relevant objects, extracted from the ground truth reasoning, in the model reasoning traces. Fig.~\ref{fig:reasoning_eval}  (middle) reports both object-level recall, where we use the complete description of an object (\eg `left half of an apple'), and noun-level recall where we use only the main noun in the object description (\eg `apple'). All four models are only able to reference 20-50\% of the relevant objects mentioned in the ground truth reasoning traces which could explain the low scores for Perceptual correctness. This also provides a clear indication that models may not be focusing on the correct objects in the videos, motivating our method in Sec.\ref{sec:visual-annotations}, which highlights the key objects in space and time and feeds them to the model at test time.

\subsubsection{Frame Ablations.}\label{sec:ablations}

Tab.~\ref{tab:frame_ablation} analyzes the influence of how many video frames are sent with the question. We limit this ablation to Gemini due to costs. As can be seen, both Gemini models continue to benefit from an increased number of frames. This demonstrates that our dataset requires fine-grained temporal information to solve, and that Gemini can effectively utilize large contexts. Note existing reasoning datasets such as MINERVA~\cite{nagrani2025minerva} tend to plateau at around 256 frames.  On the other side of the spectrum, given only the question and answer choices and no visual data (\ie frames), performance is barely above random, with single-frame performance not much better showing that temporal information is necessary.

\begin{table}[]
\centering
    \caption{{\bf Ablation of \# frames.} Visual information is necessary to solve the dataset.
    Providing more frames improves results.
    }
    \label{tab:frame_ablation}
    \centering
    \resizebox{.9\linewidth}{!}{
    \begin{tabular}{lcc}
    \toprule
         {\bf \# frames} & {\bf Gemini 2.5 Flash} & {\bf Gemini 2.5 Pro} \\
         & No thinking & \\
         \midrule
         0 (QA-only, no visual data) & 24.5 & 27.5 \\
         1 & 23.0 & 29.3 \\
         64 & 31.7 & 40.1 \\
         256 & 37.7 & 44.5 \\
         512 & 40.1 & 49.1 \\
         1024 & \textbf{40.8} & \textbf{49.8} \\
         \bottomrule
    \end{tabular}
    }  
\end{table}

\begin{figure*}[ht]
\centering
\includegraphics[width=\linewidth,trim={11.5mm 0 56.5mm 0},clip]{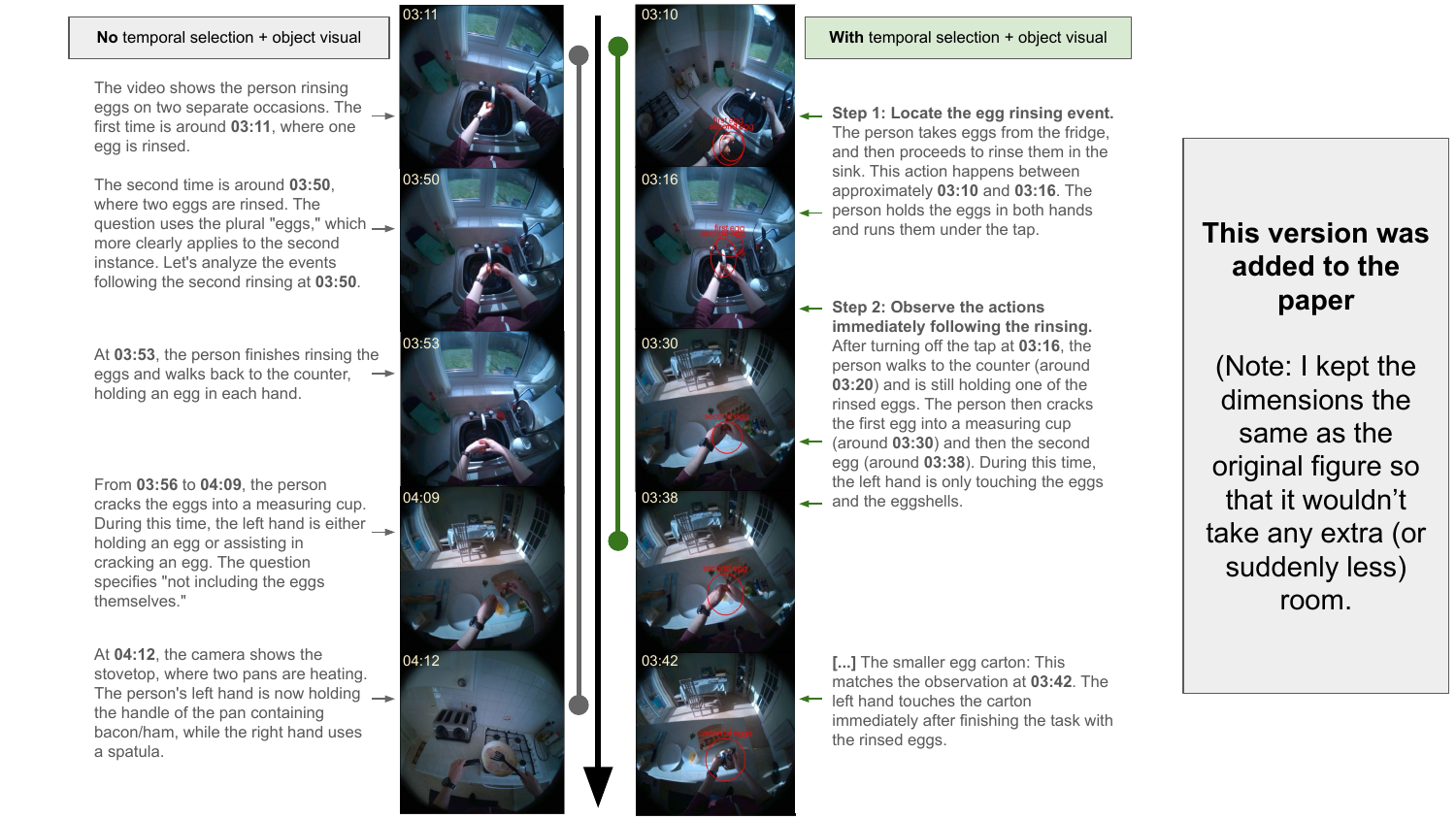}
\caption{Visualization of the reasoning traces without (Left) and with (Right) spatio-temporal hints
for the question ``\textit{After rinsing the eggs, what is the first item, not including the eggs themselves, that I touch with my left hand?}'' where the correct answer is ``\textit{Smaller egg carton}''. On the left, without spatio-temporal hints 
the model fails to identify the correct temporal span because it misses one of the two eggs at 03:11, which makes it focus on the wrong time noting the handle of the pan at 04:12. Instead the visualizations make it clear there are two eggs which focuses the model to the correct time where it is helped more by highlighting the egg carton at 03:42. Note that time stamps are only rendered for reference; only the red annotations are present for model input visualization. Best viewed zoomed in and in colour.
}
\label{fig:reasoning_vis}
\end{figure*}

\subsection{Spatio-temporal Highlighting}
\label{sec:visual-annotations}


Recent works such as Set-of-Mark~\cite{yang2023setofmark} show that LVLMs significantly benefit from highlighting important objects in the visual domain by directly altering the visual input (i.e. pixels). Therefore we explore whether segmentation can help via a variety of visual or textual aids. We do this first in an oracle setting and then with real segmentation results.

\begin{figure*}[th]
\vspace{-0.7cm}
    \centering
    \includegraphics[width=0.25\linewidth]{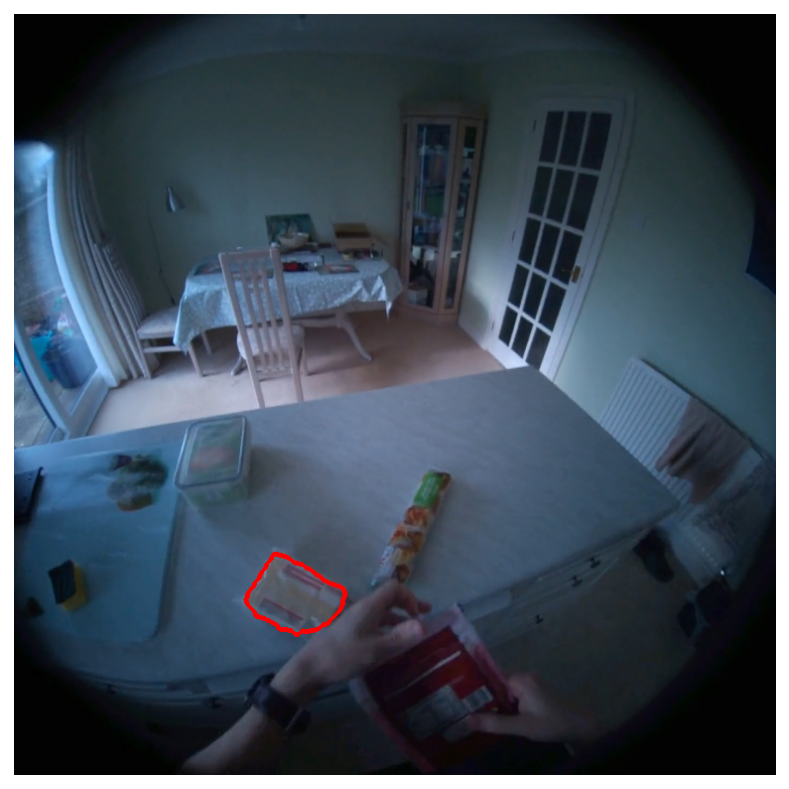}
    \includegraphics[width=0.25\linewidth]{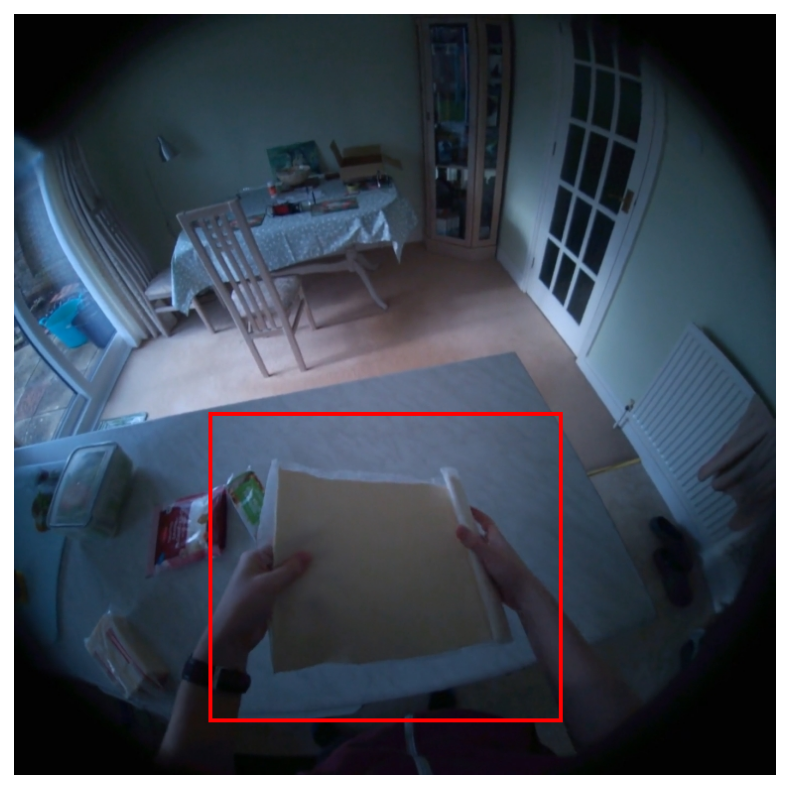}
    \includegraphics[width=0.25\linewidth]{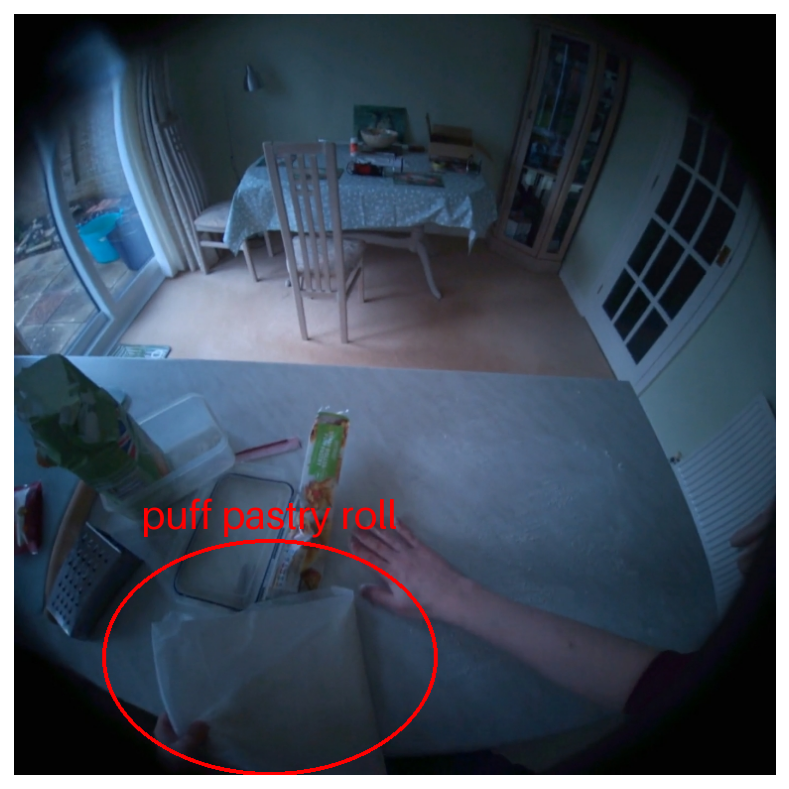}
    \caption{{\bf Visual annotations.} From left to right, visualization using \emph{masks}, \emph{boxes}, and \emph{circles + classes} (See Tab.~\ref{tab:visual_annotations}). The target question is, ``How many types of cheese did I add to the puff pastry roll?''}
    \label{fig:visual_annotations}
    \vspace{-.2cm}
\end{figure*}

\para{Oracle experiment.}
The original HD-EPIC~\cite{perrett2025hdepic} dataset comes with some ground-truth segmentation annotations: each time an object is moved, the object is segmented in the first and last frame of this movement. Since this is quite sparse, we use our own re-implementation of SAM2~\cite{ravi2024sam2} to fill in the segments of the frames in between. Hence we have decent segmented tracks of all objects when they are moved.
For each object mentioned in the reasoning traces,
if it was annotated, we linked to the corresponding object ID. This link exists for 82\% of the objects in our reasoning traces.
Using this, we can highlight for each question the (majority of) relevant objects. We highlight the object by drawing a mask contour in red, or surrounding the object by a box or ellipsoid.
Additionally, we draw the class name in the frames. Fig.~\ref{fig:visual_annotations} visualizes what these annotations look like. Furthermore, since Gemini natively predicts class labels and boxes, instead of highlighting the objects we interleave frames with objects and box coordinates using the Gemini box prediction format (non-json). 


Next, since the objects are localized in time, we can use this to select important frames to feed to the LVLM. More specifically, we take all frames where the objects relevant to the question are present, subsampling if necessary. If there is remaining frame budget, we uniformly sample the rest.

\begin{table}[]
    \caption{{\bf Ablation with different visualization hints - oracle.} We use Gemini 2.5 Pro with 256 frames and highlight relevant objects in various ways. We do this with and without temporal selection. Circles, classes, and temporal selection all help and can be combined to yield an improvement of 5.6\%. 
    Interleaving frames with text-based box coordinates and class names works less well than our highlighting approach. 
    }
    \label{tab:visual_annotations}
    \centering
    \resizebox{\columnwidth}{!}{
    \begin{tabular}{lcc}
    \toprule
         {\bf visualization} & {\bf without} & {\bf with} \\
         {\bf type} & {\bf temporal selection} & {\bf temporal selection} \\
    \midrule
         none (baseline)     & 44.5\%  & 47.4\% \\
         masks               & 44.8\%  & 47.3\% \\
         boxes               & 45.1\%  & 47.6\% \\
         circles             & 45.9\%  & 48.6\% \\
         classes             & 47.1\%  & 49.2\% \\
         circles and classes & 47.3\%  & \textbf{50.3\%} \\
    \midrule
         boxes and classes as text & 46.6\% & 48.4\% \\ 
    \bottomrule
    \end{tabular}
    }  
\end{table}
\begin{table}[t]
    \centering
    \resizebox{\columnwidth}{!}{
    \begin{tabular}{lcc}
    \toprule
         {\bf visualization} & {\bf without} & {\bf with} \\
         {\bf type} & {\bf temporal selection} & {\bf temporal selection} \\
    \midrule
         none (baseline)     & 44.5\%  & 46.5\% \\
         circles             & 44.0\%  & 43.0\% \\
         classes             & 44.8\%  & \textbf{47.0\%} \\
         circles and classes & 44.1\%  & 44.2\% \\
    \midrule
         boxes and classes as text & 43.7\% & 45.5\% \\
    \bottomrule
    \end{tabular}
    }  
    \caption{{\bf Ablation with different visualization hints - predicted objects.} We use Gemini 2.5 Pro with 256 frames to ask what objects would be useful to highlight to answer the question. We detect these with OWLv2~\cite{minderer23neurips_owlv2}. We then highlight the detected objects with and without temporal selection. Like the oracle experiment, temporal selection and specifying class names improves results by up to 2.5\%. Unlike the oracle experiments, highlighting the objects with circles does not seem to help.}
    \label{tab:real_visual_annotations}
\end{table}

\begin{table}[t]
    \centering
    \small{
    \begin{tabular}{lc}
    \toprule
   \textbf{ Method} & \textbf{Gemini acc.} \\
    \midrule 
    baseline (256 frames) & 44.5\%\\
       \quad + spatial hints & 44.8\%\\
    \quad \quad + temporal hints  & \textbf{47.0\%} \\
    \midrule 
    \quad + spatial hints (\textbf{oracle}) & 47.3\%\\
    \quad \quad + temporal hints (\textbf{oracle})  & \textbf{50.3\%} \\
    \bottomrule
    \end{tabular}}
    \caption{\textbf{Spatio-temporal highlighting on \dataset{}} We show results for both the oracle case and our agentic spatio-temporal highlighting method Gemini 2.5 Pro.}
    \label{tab:spatial_highlighting}
    \vspace{-.3cm}
\end{table}


Results in Tab.~\ref{tab:visual_annotations} show that encircling the object is a better way to highlight than drawing masks or bounding boxes, leading to 1.2-1.4\% accuracy improvements. This is likely because encircling is typically a more natural form of highlighting (outside our community) than drawing boxes, while we observed that masks are sometimes noisy which results in unnatural visualizations. Writing the class names yields good improvements of 2.6-2.8\% over no visualizations. Just doing temporal selection on the baseline yields an improvement of 2.9\%. These three techniques are orthogonal and their combination yields 50.3\% accuracy, an improvement of 5.8\%. Interleaving boxes with classes as text between frames performs worse than highlighting them in the frames, likely because the bounding box text format seems mostly related to individual images. Fig.~\ref{fig:reasoning_vis} shows a qualitative example where highlighting the relevant objects in the frame helps the model identify the correct temporal span which leads to the correct reasoning and final answer. 

\begin{figure*}
\vspace{-.7cm}
\hspace{-1.0cm}
 \begin{tabular}{>{\RaggedRight}m{0.23\linewidth}>{\RaggedRight}m{0.55\linewidth}>{\RaggedRight}m{0.25\linewidth}}

    \multicolumn{1}{c}{\textbf{Question}} & 
    \multicolumn{1}{c}{\textbf{Model Reasoning}} & 
    \multicolumn{1}{c}{\textbf{Ground Truth Answer}} \\
     \hline

    
    
    \scriptsize
    In what order do I drink water, put the second knife into the dishwasher, and pour hot water into the flask? Possible answer choices:
    
    (1) Put the second knife into the dishwasher, drink water, pour hot water into the flask
    
    (2) Drink water, put the second knife into the dishwasher, pour hot water
    
    (3) Pour hot water into the flask, drink water, put the second knife into the dishwasher
    
    (4) Put the second knife into the dishwasher, pour hot water into the flask, drink water
    
    & 
    \scriptsize
    \textbf{Gemini 2.5 Pro}
    
    The correct answer is (4). Here's why:

 \textcolor{red}{1. **Drinking water:** The video doesn't explicitly show the person drinking water.}
2. **Pour hot water into the flask:** Starting at image 157, the person is seen filling a metal flask with hot water from a kettle. This action is completed by image 162.
3. **Put the second knife into the dishwasher:** In images 152-155, the person is placing knives in the dishwasher. After pouring water into the flask, in images 164-167 the person puts another knife into the dishwasher.

\textcolor{red}{Therefore, the order of events that are shown in the video is putting the second knife into the dishwasher, pouring hot water into the flask, then potentially drinking water after.}
    
    & 
    \scriptsize
    1. At 00:59, I notice the user putting the first knife ('knife-2c3afc468fb10cb2') into the dishwasher.

2. At 03:06, I notice the user drinking water.

3. At 04:39, I notice the user pouring hot water into the flask ('flask-9fb1d013fbbfb9ed').

4. At 05:50, I notice the user putting the second knife ('knife2-067f49caca9d1aff') into the dishwasher.

\textcolor{green!50!black}{5. The correct order of activities, therefore, is drinking water, pouring hot water into the flask, and putting the second knife into the dishwasher.}
    \\
    \hline

    \scriptsize
    How many lemons did the user shave with the grater? 
    
    Possible answer choices:
    
    (1) 0 
    
    (2) 2
    
    (3) 3 
    
    (4) 1 
    
    (5) 4

     & 
    \scriptsize
    \textbf{Claude Sonnet 4}

     I need to carefully look through the video frames and count how many lemons the user shaves/grates with a grater.
Looking through the frames, I can see:

- At \textcolor{red}{05:24 and 05:48}: The user appears to be handling what looks like a lemon (yellow object) and using a grater/grating tool ...
Let me look more carefully at the grating activity:

- Around 05:24, 05:48: I can see the person holding what appears to be a yellow lemon and using a grater. It looks like they're grating one lemon.

- \textcolor{red}{I don't see clear evidence of multiple lemons being grated in sequence in the frames provided.}...


\textcolor{red}{From what I can observe in the frames, it appears the user grated 1 lemon with the grater. I don't see evidence of multiple lemons being grated.Looking through the video frames, I can see the user grating what appears to be a yellow lemon around the 5:24 and 5:48 timestamps. From the frames provided, I can observe the user grating one lemon with the grater.}
\textcolor{red}{Answer: (4)}

& 
    \scriptsize
    1. I watch the video and first notice the user begin shaving a lemon (lemon-b00bb617003dacac) with the grater at \textcolor{green!50!black}{02:57}. This is the first lemon.
    
2. I continue watching the video and notice the user begins to shave a second lemon (lemon2-56e4b71d5bfca0e3) at \textcolor{green!50!black}{04:17}.

3. Continuing to watch the video I then notice the user put the grater away at 06:06 as well as the rest of the lemons at 06:40 and conclude the total number of lemons grated is two.

\textcolor{green!50!black}{Answer: (2)} 

   \\
    \hline

\scriptsize
    Why did the green and white mat move?
    Possible answer choices:
    
    (1) The user kicked it away
    
    (2) The mat did not move
    
    (3) The vacuum pulled it
    
    (4) The user picked it up
    
    (5) There was no mat

    & 
    \scriptsize
    \textbf{GPT-5}
    
    *Analyzing mat movement**

I need to analyze why the green and white mat moved. There are a few possibilities, like if the user kicked it, it didn't move at all, or if the vacuum pulled it, among others. In the kitchen/dining area, the mat at the sliding door has a "HELLO" design and appears to move during vacuuming. I see frames where the vacuum head gets close to the mat, suggesting it was likely the vacuum that caused the movement.**Investigating mat movement**

I’m looking closely at the frames. At 00:50, the vacuum head is near the mat, and at 00:55, it’s turning around. \textcolor{red}{It seems likely that the vacuum’s suction or pushing caused the mat to move. I’m not seeing any frame where someone lifts the mat or kicks it.} Early frames show the mat closer to the sliding door, but by 00:50 it appears shifted. \textcolor{red}{So, I’m confident in choosing option (3): the vacuum pulled it.(3)}

 & 
 \scriptsize
 1. I watched the video and found a green and white mat (mat-eba912ec072326a3) at 00:01.
2. \textcolor{green!50!black}{I saw the green and white mat (mat-eba912ec072326a3) move at 00:48.}
3. Just before the green and white mat (mat-eba912ec072326a3) moved at 00:48, \textcolor{green!50!black}{I saw the user touch and pick it up.}
4. Therefore, the green and white mat (mat-eba912ec072326a3) moved at 00:48, because the user picked it up.

\textcolor{green!50!black}{Answer: (4)} 
\\ \hline

\end{tabular}
  \caption{\textbf{Example errors made by the models in their video understanding and reasoning.} The errors include failing to identify actions due to the egocentric viewpoint, missing the relevant temporal spans, and failing to capture actions due to fast motion.}
  \vspace{-2mm}
  \label{fig:dataset-examples}
\end{figure*}

\para{Using a segmentation model.}
We now repeat the experiment using the predictions of an open-vocabulary segmentation model, OWLv2~\cite{minderer23neurips_owlv2}, thus using no ground truth knowledge.
However, this model requires input classes, which we obtain by feeding the video with the original \dataset{} question (but without the reasoning trace) to Gemini 2.5 Pro and ask it to ``Identify the object categories which would be useful to identify in the video to help answer the question correctly.'' To focus on interesting objects, we remove the `hand' and `person' category as well as large static objects which in HD-EPIC~\cite{perrett2025hdepic} are called \emph{fixtures} (\eg cupboard, drawer), \emph{storage space} (\eg shelves, hooks) and \emph{large appliances} (\eg fridge, microwave). In preliminary experiments, we found that using a slightly conservative detection threshold of 0.6 worked best. In effect, this resulted in detected objects in 26\% of the frames. In frames with an object, there are on average 1.26 objects.

\begin{table}[t]
    \centering
    \small{
    \begin{tabular}{lc}
    \toprule
   \textbf{ Method} & \textbf{Gemini acc.} \\
    \midrule 
    baseline (32 frames) & 39.9\% \\
       \quad + spatial hints & 41.1\%\\
    \quad \quad + temporal hints & \textbf{ 42.1\% } \\
    \bottomrule
    \end{tabular}}
    \caption{\textbf{Spatio-temporal highlighting on  EgoTempo~\cite{plizzari2024egptempo}.} We show results using our agentic spatio-temporal highlighting method for Gemini 2.5 Pro. We cannot show oracle results as EgoTempo does not have ground truth spatio-temporal reasoning traces.}
    \label{tab:egotempo}
    \vspace{-.3cm}
\end{table}

Using the detected segments, we use circles and classes for visualizations as they were found to be the best in the oracle experiment. Results are shown in Tab.~\ref{tab:real_visual_annotations}. Similar to the oracle experiment, temporal selection improves results from 44.5\% to 46.5\%; an increase of 2.0\%. If we also visualize the class names in the frames, we obtain a slightly higher accuracy of 47.0\%. Interestingly, highlighting objects with circles does not seem to help in this setting. But again, interleaving boxes and classes as text in-between frames performs worse than visual highlighting.

We apply our best settings from the above ablations and show results in Table \ref{tab:spatial_highlighting}. We show that hinting the model to focus on relevant objects and frames can bring performance improvements in video question answering.  However, there remains a gap between oracle hints derived from the ground truth and hints from an off-the-shelf segmentation model run at inference time, showing a gap in the ability of LVLMs to attend to the objects and frames that are relevant to answer questions about videos.

\para{Generalization to EgoTempo~\cite{plizzari2024egptempo}}
To verify that our highlighting technique also generalizes to other datasets, we apply a preliminary study on the EgoTempo dataset~\cite{plizzari2024egptempo}, which is also egocentric and involves object manipulation. Again, we obtain useful object categories by feeding the video with the question to Gemini. Then we apply the same object detector as before, annotate the videos, and feed these to Gemini 2.5 pro. Since videos are much shorter, and since we want to explore results with temporal selection, we extract frames at 1fps but send only 32 frames. To verify whether the answer is correct we use Gemini 2.5 Flash as a judge.

Results in Tab.~\ref{tab:egotempo} demonstrate similar trends as before. Adding spatial hints improves acc. from 39.9\% to 41.1\%. Temporal selection improves this further to 42.1\%, showing that spatio-temporal highlighting can lead to improvements in performance, even with a small 32 frame budget.

\subsection{Qualitative Analysis}
\label{sec:qualitative-analysis}

A few examples of video, question and reasoning traces are shown in Figure~\ref{fig:dataset-examples}. Note how the reasoning traces can be very informative in identifying weak spots in model capabilities - errors include failing to identify actions due to the egocentric viewpoint, missing the relevant time region, and failing to capture fine-grained actions due to fast motion.

\section{Conclusion}

We introduce \dataset{}, 
a benchmark for evaluating complex egocentric visual reasoning consisting of a set of challenging, multi-step multimodal questions and spatiotemporally-dense human-annotated reasoning traces. Through extensive evaluations, we identify surprising successes and key areas for improvement for frontier models in video reasoning. We leverage our fine-grained annotations to better categorize reasoning errors at a more nuanced level, and show the value of our spatio-temporal highlighting for improving performance.

{
    \small
    \bibliographystyle{ieeenat_fullname}
    \bibliography{main}
}
\clearpage
\setcounter{page}{1}
\maketitlesupplementary
\section{\dataset{}}
\subsection{Additional Statistics}
\label{sec:additiona-stats}
\begin{figure}[t]
\centering
  \includegraphics[width=0.9\linewidth]{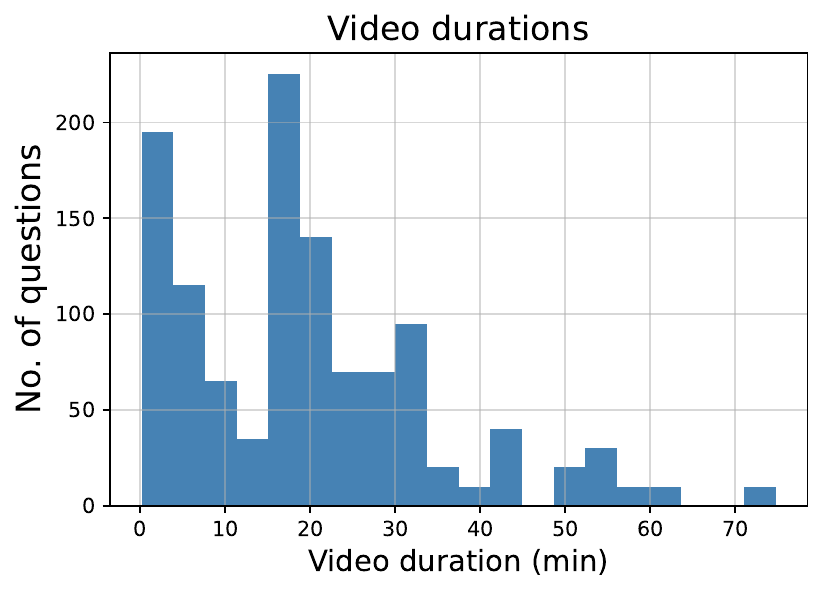}
  \caption{\textbf{Video Lengths in \dataset{}.}}
  \label{fig:video-lens}
\end{figure}

The shortest videos is 10 seconds while the longest video is 75 minutes.
The mean video length is 20 minutes. The full distribution can be found in Fig. \ref{fig:video-lens}.

\begin{figure}[t]
 \centering
 \includegraphics[width=\linewidth]{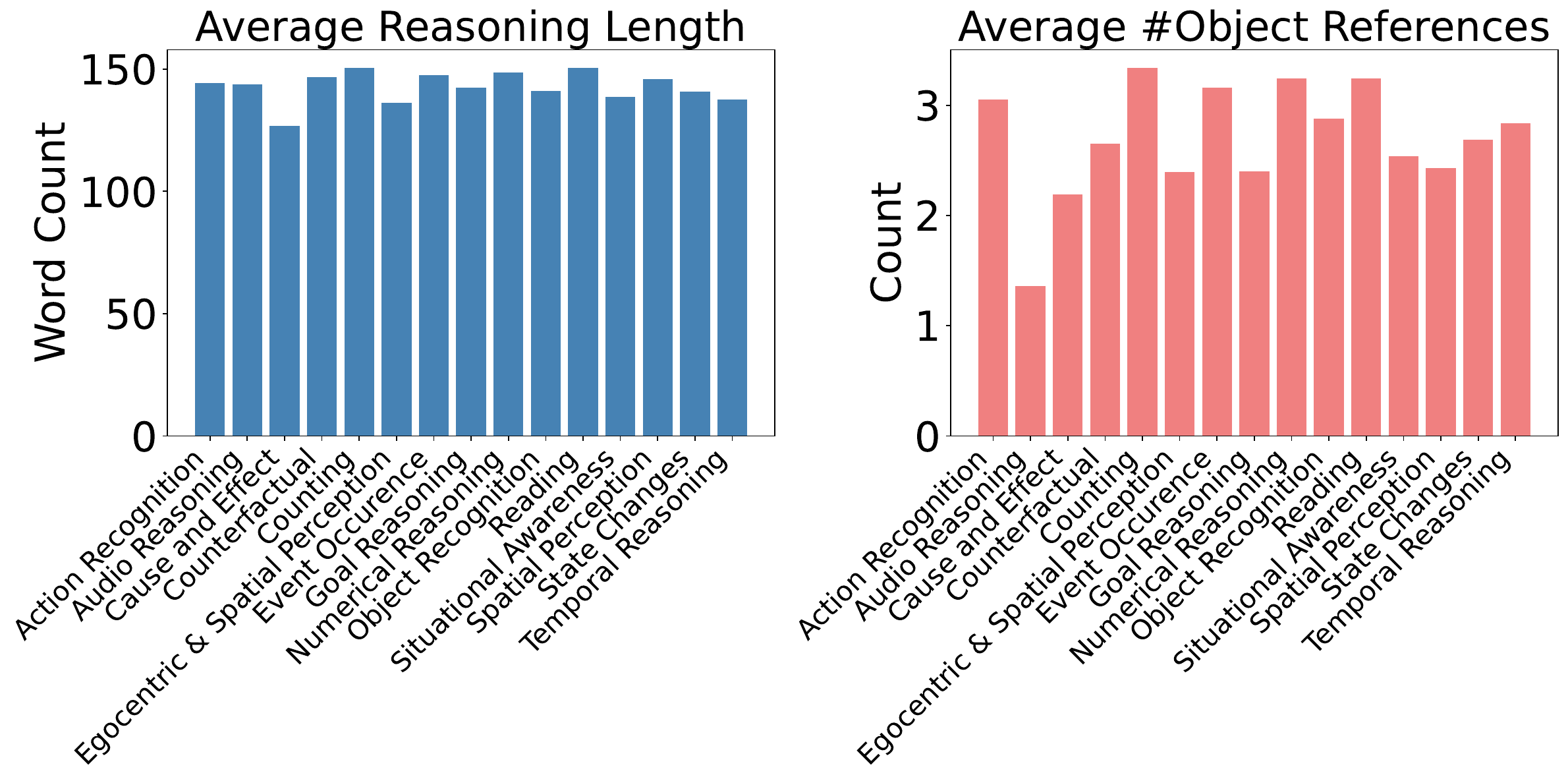}
  \caption{Statistics of reasoning trace lengths and object references broken down by specific question types.}
  \label{fig:question-type-stats}
\end{figure}

Figure~\ref{fig:question-type-stats} shows statistics of reasoning lengths and object occurrences by question type. We note that the distribution is flat, other than \textit{Audio Reasoning}, which has fewer object references (as expected).

\subsection{Dataset Post-processing details}
\label{sec:dataset-creation-appendix}
As described in Sec.~\ref{sec:dataset-creation}, our pipeline consists of selection, annotation + verification, and post-processing. We elaborate here on the verification and post-processing stages.
We follow best practices and guidelines that have been established in prior work \cite{nagrani2025minerva,hsieh2023sugarcrepe,le2020adversarial,wang2024lvbench}, which note that benchmarks often have issues regarding modality biases (\eg questions that are answerable with only text), and propose solutions to help mitigate these. The verification stage is described in the main paper, and was conducted through manual human studies, separated reviewer pools, and several rounds of intervention and refinement. In this context, the post-processing stage is a further step of refinement on top of our initial (model-free) process. Here, we use the same model ensemble of open and closed-source text-only language models in \cite{nagrani2025minerva} (\eg Deepseek, GPT-4o, Gemini Flash, and Qwen) and take their \textit{consensus} agreement as a set of candidate questions that merit further review. As described in prior work, this technique is designed in order to avoid discarding difficult questions that models may have answered correctly by chance. Notably however, rather than completely discarding all questions in the consensus set like in prior work, this set was subject to one more round of review -- if the question could be ``corrected'' (\eg the text bias was readily clear and it was clear how a change in the question or answer choices could remove this bias) then this measure was taken (we found that this was the case in the majority of the questions, where ``too much'' was given away in the question itself). If the question required more than a simple change, the question was removed.



\section{Rater Guidelines}
All textual data in \dataset{} was manually annotated by human annotators (raters). All raters are native English speakers with graduate degrees.
\subsection{Annotation Guidelines}
The raters were given the following guidelines before being asked to propose question, answers, decoys and reasoning traces: \\
You will be given a video. For each video, we would like you to propose complex multi-step questions. Questions should be accompanied by 5 answer candidates, one correct answer, and a detailed reasoning trace. The reasoning trace should refer to objects present in the video (more details are provided below). 

Questions should require 2 or more of the skills provided below: 
\begin{enumerate}
    \item Temporal Reasoning
    \item Counting
    \item Cause and Effect 
    \item Goal Reasoning
    \item Egocentric and Spatial Perception
    \item Event Occurrence
    \item State Changes
    \item Reading (OCR)  
    \item Numerical Reasoning (all math operations other than counting)
    \item Object Recognition
    \item Action Recognition
    \item Counterfactual (“what if”)
\end{enumerate}
Important things to keep in mind: 
\begin{enumerate}
    \item Questions should be multi-step
    \item 
They should be difficult to solve 
\item 
Ideally they should involve looking at multiple different time segments of the video 
\item 
Each question should be cover multiple skills and have multiple reasoning steps
\item 
The questions should be phrased with the word “I”, as if the cameraperson is asking the question.
\item 
The reasoning traces should use the word “I” to refer to you (the rater, as you are doing the reasoning, and use the word “user” to refer to the cameraperson). Note in the question, I refers to the camera person. 
\item 
The reasoning traces should map to the “objects of interest”.
\item 
Every time that object is referenced (even if it is in a different time step of the video), please use the same annotation.
\item 
Please number the “steps” in each reasoning trace, as shown in the examples.
\item 
All reasoning traces must have at least one time stamp, and at least one object of interest reference.
\end{enumerate}

\subsection{Human Study}
\label{appendix:human_study_guidelines}
We used a disjoint pool of 12 raters for the human study, by asking the rater leads to ensure that the same rater who proposed the question (or even saw the video before) is not the same as the one who performs the human study. Hence no rater performing the humans study has seen the video before. All raters had access to the full videos. Rater guidelines are below: 

Goal: You will be given a video, a question, and 5 answer choices. Please watch the video and pick the correct answer (A, B, C, D, E). (Like a multiple choice exam). You can watch the video as many times as you like, and you can rewatch various parts of the video. Please take as much time as you require.
\\

\section{Model Implementation Details}
Hyperparameters for all our models are provided in Table~\ref{tab:hyperparams}. 
\begin{table*}[ht]
\centering
\begin{tabular}{llr}
\toprule
\textbf{Method}  & \textbf{\# of Frames} & \textbf{Hyperparameters (seeds, temperature, etc)}  \\
 
\midrule
Qwen-3~\citep{qwen3techreport} & all @ 1fps &  \parbox{6cm}{version=Qwen/Qwen3-VL-235B-A22B-Thinking, \\ max-sequence-length=128k} \\
\midrule
GPT-5~\citep{openai25gpt5} & 64 &  \parbox{6cm}{version=gpt-5-2025-08-07, \\ seed=default, \\ top-p=default, \\ temperature=default, \\ image resolution=auto, \\ reasoning effort=medium} \\
\midrule
{GPT-4.1~\cite{openai-gpt4.1}} & 64 &  \parbox{6cm}{version=gpt-4.1-2025-04-14, \\ seed=default, \\ top-p=default, \\ temperature=default, \\ image resolution=auto} \\
\midrule
Claude Sonnet 4~\cite{claude4opus_sonnet} & 64  &  \parbox{6cm}{version=claude-sonnet-4-20250514 \\ seed=default, \\ top-p=default, \\ temperature=1.0, \\ thinking budget=10,000 } \\ 
\midrule
{Gemini 2.5 Flash Thinking~\citep{gemini2.5}}  & 64 & \parbox{6cm}{temperature=default, \\ seed=default, \\ sampling=default} \\
\midrule
{Gemini 2.5 Pro Thinking~\citep{gemini2.5}}  & 64 &  \parbox{6cm}{temperature=default, \\ seed=default, \\ sampling=default} \\
\bottomrule
\end{tabular}
\caption{\textbf{Hyperparameters for all model baselines}}
\label{tab:hyperparams} 
\end{table*}
\section{Results}

\subsection{MiRA Reasoning Evaluation}
\label{sec:mira-gpt}
MiRA is a model-based metric that uses Gemini 2.5 Pro for scoring.
The exact prompt used to compute the MiRA score appears in Figure~\ref{fig:mira_prompt}.

\begin{figure*}
    \centering
\begin{exampleprompt}
You are an expert at grading student answers to questions about videos. For each video, you will get a question about the video, the correct reasoning, and the final answer. You will then get the reasoning from the student, and a set of criteria. Given this criteria, please provide a score from 0, 1 or 2 for each criterion that will assess the student's work.\\

**Criteria:**\\
(1) Perceptual correctness: was the relevant information perceived accurately from the video?\\
(2) Temporal grounding: were time ranges provided for each piece of information from the video, and if so were they accurate?\\
(3) Logical reasoning: was the reasoning logically sound, given the information perceived (independent of whether that information was correct)?\\
(4) Completeness: were any steps skipped in the given answer or left unstated?\\

**Question:** \emph{\{question\}}\\

**Reference Reasoning:** \emph{\{reference\_reasoning\}}\\

**Reference Answer:** \emph{\{reference\_final\_answer\}}\\

**Student Reasoning:** \emph{\{model\_reasoning\}}\\

For each of the above criteria, please provide a score from 0,1,2.\\
0 - Doesn’t fulfill the criteria at all\\
1 - Partially fulfills the criteria\\
2 - Completely fulfills the criteria\\
Please produce the score in the format [Criteria: score]\\

Examples:\\

**Question:** How many times heavier is the second density cube than the first? 2.64., 3.05., 2.86., 1.35., 1.77.,\\
**Reference Answer:** I began watching the video and noticed many duplicate items. At 01:57, the narrator says “density cubes,” and then shows the cubes. At 02:01, they are all labeled. There are a total of 12 density cubes shown. At 02:07, the narrator takes 2 away to weigh them. The first cube displays a number of 44 on the scale, shown at 02:11. At 02:13, the second cube is weighed. The number 116 appears on the scale. These two numbers are not said out loud. If we divide 116 by 44 and round to 2 decimal places, we get 2.64. Final Answer: 2.64.\\
**Student Reasoning**\\
The question asks how many times heavier the second density cube is than the first.\\
From [01:58] to [02:04], the video shows 12 density cubes.\\
From [02:00] to [02:04], the video labels the first cube as "acetate" and the second cube as "polypropyline".\\
From [02:11] to [02:13], the video shows the presenter putting the acetate cube on a scale, which reads 44 grams.\\
From [02:11] to [02:13], the video shows the presenter putting the polypropyline cube on a scale, which reads 24 grams.\\
To find how many times heavier the second cube is than the first, we divide the weight of the second cube by the weight of the first cube: 44/24 = 1.83.\\
The closest answer choice to 1.83 is 1.77.\\

Output should be:\\
(1) Perceptual correctness: 1\\
(2) Temporal grounding: 2\\
(3) Logical reasoning: 1\\
(4) Completeness: 2\\

**Question:** How many passes did the white team make right before they scored their first goal? 4., 1., 2., 5., 3.,\\
**Reference Answer:** I searched for the white team's first score, which happened at 00:42. I watched the clip that showed the entire play, which is from 00:31 to 00:43. I counted one pass at 00:39 and one pass at 00:41. The player who receives the pass at 00:41 shoots into the goal. To confirm, I watched the replay from 00:44 to 00:47. At 00:46, the puck bounces off a white player's stick and into the goal. This makes the shot taken at 00:41 a pass and assist instead of a proper shot, which means the white team made 3 passes before they scored their first goal. Final Answer: 3.\\
**Student Reasoning**\\
The video shows the white team making a series of passes before scoring their first goal. The sequence starts with a player receiving the puck near the blue line, then passing it to a teammate who skates towards the goal. This teammate then makes another pass to a third player who is positioned closer to the net. Finally, this third player takes a shot and scores. This sequence involves three distinct passes, which aligns with option (E). The timestamps and visual cues in the video confirm these actions, showing clear transitions between players handling the puck.\\

Output should be:\\
(1) Perceptual correctness: 2\\
(2) Temporal grounding: 0\\
(3) Logical reasoning: 2\\
(4) Completeness: 2\\

**Re-iterating the Criteria:**\\
(1) Perceptual correctness: was the relevant information perceived accurately from the video? (eg were the correct objects identified, was the text read properly from the screen, were the relevant events and actions mentioned correctly)\\
(2) Temporal grounding: were time ranges provided for each piece of information from the video, and if so were they accurate?\\
(3) Logical reasoning: was the reasoning logically sound, given the information perceived (independent of whether that information was correct)?\\
(4) Completeness: were any steps skipped in the given answer or left unstated?\\
\end{exampleprompt}
    \caption{The MiRA rubric prompt, as provided to Gemini (or GPT), used to score the reasoning traces produced by the VLMs being evaluated, as described in Section~\ref{sec:reasoning_analysis}.}
    \label{fig:mira_prompt}
\end{figure*}

\begin{figure}[h]
  \centering
  \includegraphics[width=0.78\linewidth]{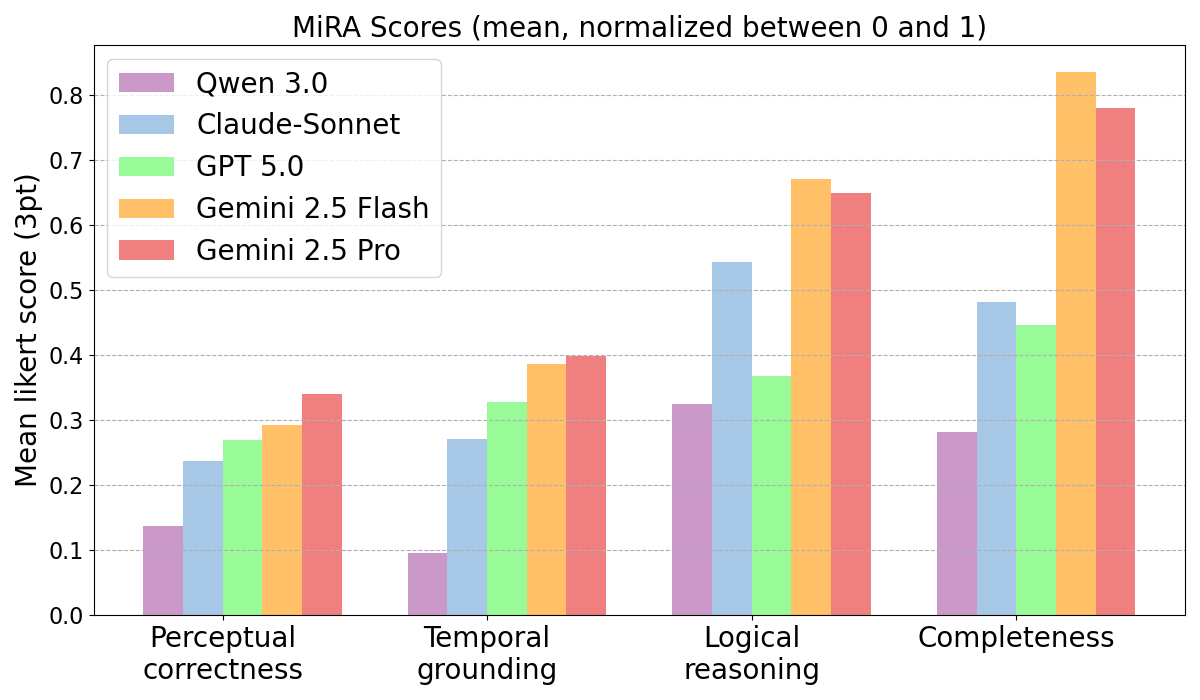}
  \caption{MiRA scores of the model reasoning traces using GPT-5 as a judge.}
  \label{fig:gpt5}
\end{figure}

Since using the same LLM for hinting and as a judge could \emph{potentially} bias the results, we repeat the MiRA analysis from Fig.~\ref{fig:reasoning_eval} (left) using GPT-5 as a judge, and report the scores in Figure~\ref{fig:gpt5}. Comparing the two, we can see that the \textit{relative ordering} of the models is preserved. Since this evaluation compares the ground truth reasoning trace with the model's using a fixed rubric we believe there is less dependence on the judge itself. 

It's worth noting that \dataset{} includes multiple-choice answers for every question, enabling MCQ accuracy to serve as a primary, model-free metric.

\subsection{Results by Skill/Length}
%
%
\begin{figure}
    \centering
    \includegraphics[width=\linewidth]{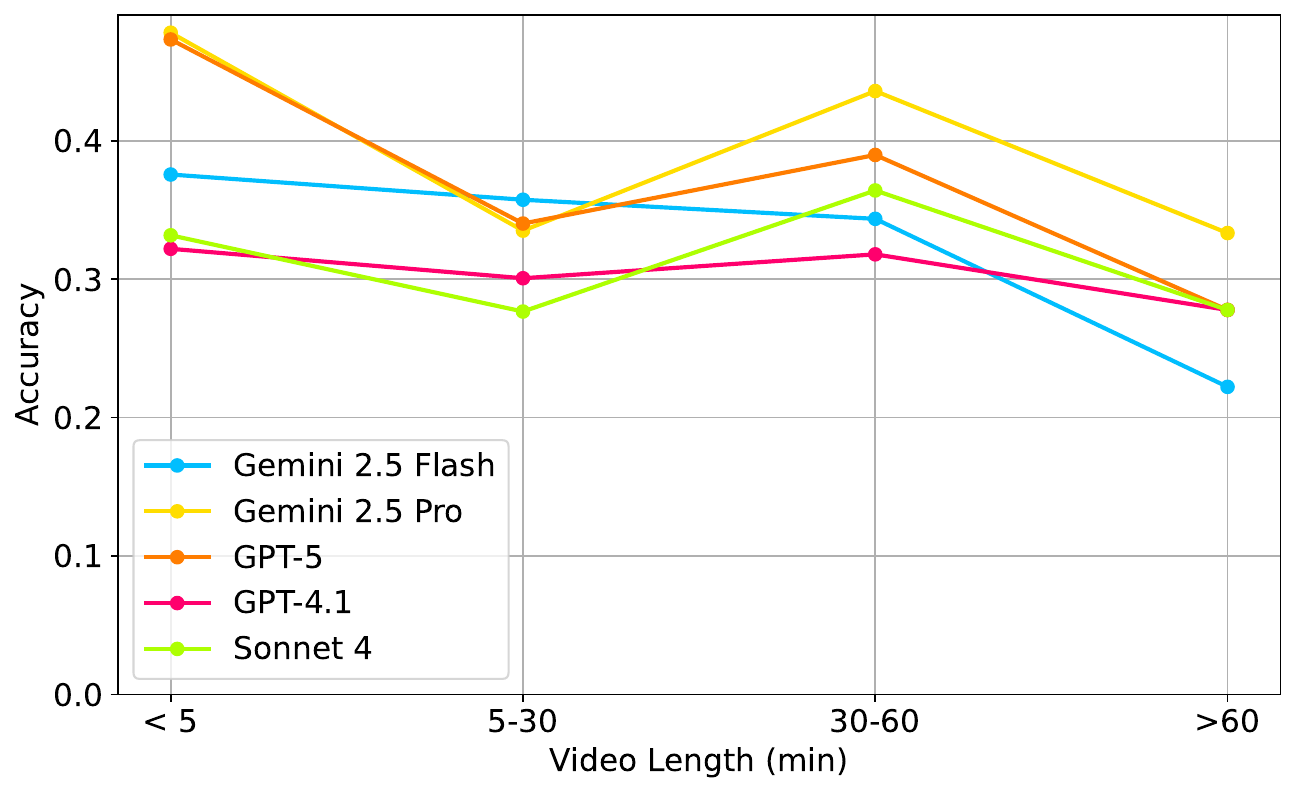}
    \caption{Accuracy for videos of different lengths}
    \label{fig:per_video_length_score_distributions}
\end{figure}
\begin{figure}
    \centering
    \includegraphics[width=\linewidth]{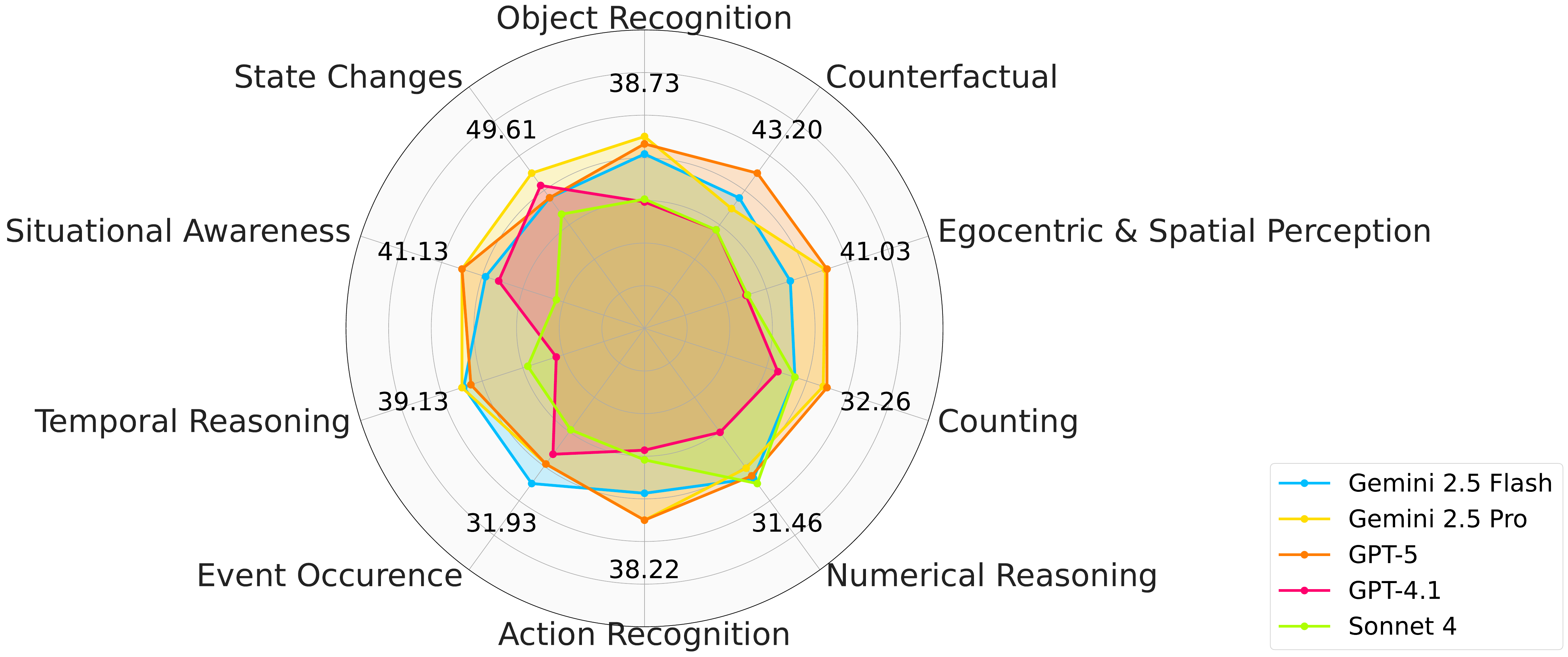}
    \caption{Accuracy of SotA models by skill}
    \label{fig:per_skill_score_distributions_radar_chart}
\end{figure}
We break down the performance of SotA models by video length in \ref{fig:per_video_length_score_distributions}. Performance of most models is remarkably robust with increasing context length, but generally declines for videos longer than one hour. \ref{fig:per_skill_score_distributions_radar_chart} breaks down performance by skills. Note that we excluded the following skills because they had less than 50 examples in the dataset: Reading, Cause and Effect, Goal Reasoning, Audio Reasoning and Spatial Perception. Generally, models perform best at questions involving State Changes, Counterfactuals and Situational Awareness, but struggle with Event Occurrences, Numerical Reasoning and Counting.


\end{document}